\pdfoutput=1

\documentclass[11pt]{article}

\usepackage[]{acl}

\usepackage{times}
\usepackage{latexsym}
\usepackage{amsmath}
\usepackage{hhline}
\usepackage{graphicx}
\usepackage{ulem}
\usepackage{bm}
\usepackage{amssymb}
\usepackage{booktabs}
\usepackage{tabularx}
\usepackage{enumitem}
\usepackage{bookmark}
\usepackage{bm}
\usepackage{amsfonts}
\usepackage{subfigure}
\usepackage{xcolor}
\usepackage{arydshln}
\usepackage{multirow}
\usepackage{stmaryrd}
\usepackage{color}
\usepackage{makecell}
\usepackage{float}

\usepackage[T1]{fontenc}

\usepackage[utf8]{inputenc}

\usepackage{microtype}

\title{EmoDM: Empathetic Response Generation with Emotion-aware \\ Dialogue Management}

\author{ Yuhan Liu\quad Jun Gao\quad Jiachen Du\quad Lanjun Zhou\quad Ruifeng Xu\thanks{\;\;Corresponding author}\\
 Harbin Institute of Technology (Shenzhen)\\
\normalsize \texttt{\{yhanliu, jgao95\}@stu.hit.edu.cn}\\ \texttt{\{jacobvan199165,bluejade.zhou\}@gmail.com}\\ \texttt{xuruifeng@hit.edu.cn}
}

\begin{document}
\maketitle
\begin{abstract}
A good empathetic dialogue system should  first track and understand a user's emotion and then reply with an appropriate emotion.
However, current approaches to this task either focus on improving the understanding of users' emotion or on proposing better responding strategies, and very few works consider both at the same time.
Our work attempts to fill this vacancy. Inspired by task-oriented dialogue systems, we propose a novel empathetic response generation model with emotion-aware dialogue management. The emotion-aware dialogue management contains two parts: (1) Emotion state tracking maintains the current emotion state of the user and (2) Empathetic dialogue policy selection predicts a target emotion and a user's intent based on the results of the emotion state tracking. The predicted information is then used to guide the generation of responses.
Experimental results show that dynamically managing different information can help the model generate more empathetic responses compared with several baselines under both automatic and human evaluations. 
\end{abstract}

\section{Introduction}
\begin{figure}[htbp!]
\centering
\includegraphics[scale=0.5]{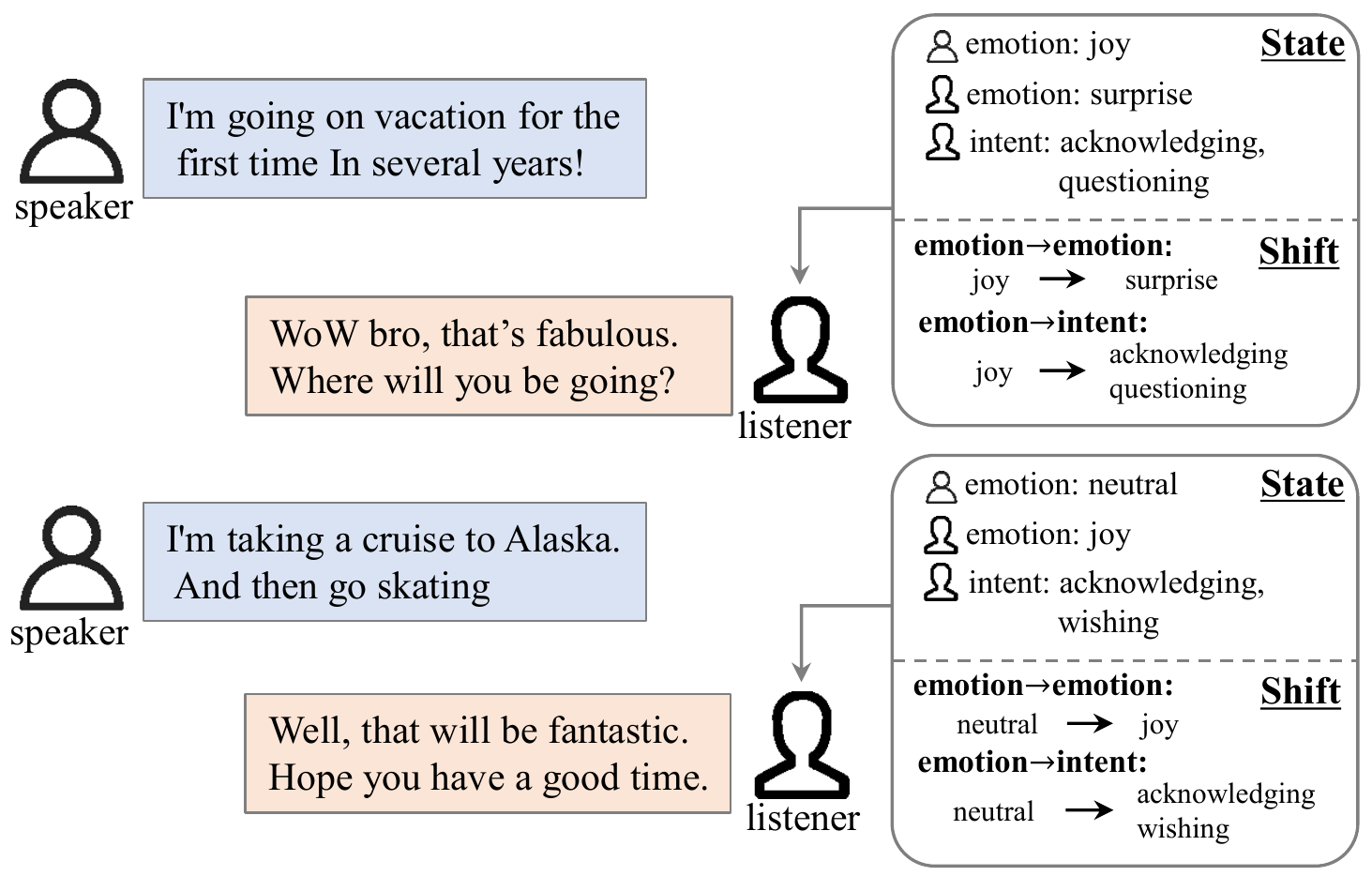}
\caption{An empathetic conversation instance with the corresponding listener states \label{fig:dialog-exp}}
\label{fig:example}
\end{figure}

Empathy plays a critical role in daily communication. 
Many researches have shown that endowing dialogue systems with empathy can improve the human-computer interaction experience \cite{Mercer2002EmpathyAQ,partala2004effects}.
In this work, we focus on empathetic response generation \cite{rashkin-etal-2019-towards}, whose goal is to enhance the ability of dialogue systems to perceive and express emotions in conversations. 
As shown with an example in Fig.~\ref{fig:example}, given a dialog context, the empathetic dialogue system (the listener) is required to track and understand the speaker's emotion state in a multi-turn dialogue scenario and reply with an appropriate emotion.

Existing approaches to empathetic response generation can be divided into two categories: the first improves understanding of dialogue emotion by tracking the user's emotion~\cite{Rashkin2018IKT,lin-etal-2019-moel,majumder-etal-2020-mime,Gao2021ImprovingER}. For example, 
\citet{majumder-etal-2020-mime} propose to generate empathetic responses by mimicking the emotion of the speaker to a varying degree, depending on its affective charge~(positive or negative) and the dialogue context. \citet{Gao2021ImprovingER} prove that recognizing the emotion cause is helpful for the empathetic response generation model to better understand human emotions.
The second improves empathetic response generation by using response emotion signals as feedback \cite{li-etal-2020-empdg} or relationships between the dialogue history and the gold responses~\cite{li2020towards}.
While promising results have been achieved by both kinds of approaches, they ignore  the interaction between the use's emotion and the responding strategy, i.e.
they either focus on improving the understanding of users' emotion or on proposing better responding strategies.
Very few works consider both at the same time, and our work attempts to fill this vacancy. 

Our work is mainly inspired by task-oriented dialogue systems, in which the dialogue management component is designed for tracking the dialogue state and selecting an executable action based on the results of the dialogue state tracing. Similarly, a good empathetic dialogue system should first track and understand the user’s emotion and then reply with an appropriate emotion based on the user’s emotion. To achieve this,  we propose a novel empathetic response generation model with emotion-aware dialogue management. 
It contains two components: (1) Emotion State Tracker that maintains the current emotion state of the user and (2) Empathetic Dialogue Policy Predictor that predicts the response emotion and the intent of the response to be generated, based on the results of the emotion state tracker. The predicted information is then used to guide the generation of responses. As shown in Figure \ref{fig:dialog-exp}, in the first turn, the listener shows a \textit{surprise} feeling in consideration of the speaker's \textit{joyful} emotion, and the listener expresses his empathy with two kinds of intents. The speaker's emotion, the listener's emotion, and the listener's intent construct a triple. We define such triple as an emotion state, and the response generation can be regarded as an emotion state shift process. 

To model interaction between the use's emotion and the responding strategy, we also introduce emotion shift patterns. 
Intuitively, there exists emotion shift patterns among \textit{Empathetic Dialogue} \cite{rashkin-etal-2019-towards}. We make a statistical experiment and have the following findings: probability that \textit{surprise} shifts to \textit{joy} is 32.2\% and probability that \textit{neutral} shifts to \textit{surprise} is 30.3\%, and human observation to the subset of the dialogue data proves that responses are influenced by the cooperation of speaker's emotion and listener's emotions \cite{zhou2018emotional} rather than the other one. Besides, \citet{welivita-pu-2020-taxonomy} provide a dataset annotated with intents, and they find  strong evidence that there are shift patterns between the speaker's emotions and response intents over multiple turns, which can provide a solution to detect goals, relationships and empathy. 



Overall, we make the following contributions:
\begin{itemize}[leftmargin=*,noitemsep,nolistsep]
    \item Inspired by advance in dialogue management from task-oriented dialogue systems, we propose an emotion-aware dialogue management method for empathetic response generation. It contains two components: the Emotion State Tracker and the Empathetic Dialogue Policy Predictor.
    \item To model interaction between the use's emotion and the responding strategy, we devise emotion shift patterns. Our preliminary experiments show that there exists emotion shift patterns among \textit{Empathetic Dialogue}.
    \item Experimental results show that our model outperforms other state-of-the-art baselines in both automatic and human evaluations, and the proposed emotion-aware dialogue management can effectively improve empathetic response generation.
\end{itemize}


\section{Related Work}
Recently, open domain neural dialogue systems \citep{Gao_Bi_Liu_Li_Shi_2019,lowe-etal-2017-towards,DBLP:TransferTransfo,zhang-etal-2020-dialogpt} have get promoted in a tremendous speed. Specifically, emotionalizing \citep{ghosh-etal-2017-affect,zhou2018emotional,zhou-wang-2018-mojitalk,song-etal-2019-generating,shen-feng-2020-cdl} and personalizing \citep{ijcai2017-521,chu-etal-2018-learning} the chatbots have become a tendency. 

With the application of deep learning technique, emotion-aware response generation gets promoted dramatically. \citet{zhou2018emotional} uses internal and external memory modules to store emotion features for explicitly generating different emotional responses. \citet{huang-etal-2018-automatic} utilizes several approaches to infuse emotions into encoders and decoders to generate emotional responses. \citet{colombo-etal-2019-affect} proposes word-level and sequence-level emotion regularizers and emotion-oriented sampling method. \citet{song-etal-2019-generating} adopts lexicon-based attention and sequence-level emotion classifier to increase emotion intensity during generation. 

Empathetic dialogue generation can be seen as a sub-task of emotion-aware response generation, whose goal is to generate responses with proper emotions. \citet{rashkin-etal-2019-towards} proposes a standard benchmark that contains large-scale multi-round empathetic conversations. Then MoEL \citep{lin-etal-2019-moel} is proposed, which uses several dedicated emotion decoders to generate softly-combined outputs. \citet{shin2020generating} applies reinforcement learning to introduce user feeling towards the generated response during training. \citet{li2020towards} leverages multi-type external knowledge and emotional signal distilling to enhance information flow. EmpDG \citep{li-etal-2020-empdg} focuses on the potential of user feedback \citep{ijcai2020-503} and introduces an adversarial learning framework to capture the nuances of user emotion. MIME \citep{majumder-etal-2020-mime} assumes that empathetic responses should mimic speakers' emotion and adds stochasticity during response generation. \citet{gao-etal-2021-improving-empathetic} utilizes emotion cause information within utterances to help dialog systems better understand human emotions for empathically responding. \citet{CAiRE2019} builds an online empathetic chatbot demo with pre-trained language model GPT2 \citep{radford2019language}.

\section{Methodology}

\begin{figure}[ht]
\centering
\includegraphics[scale=0.7]{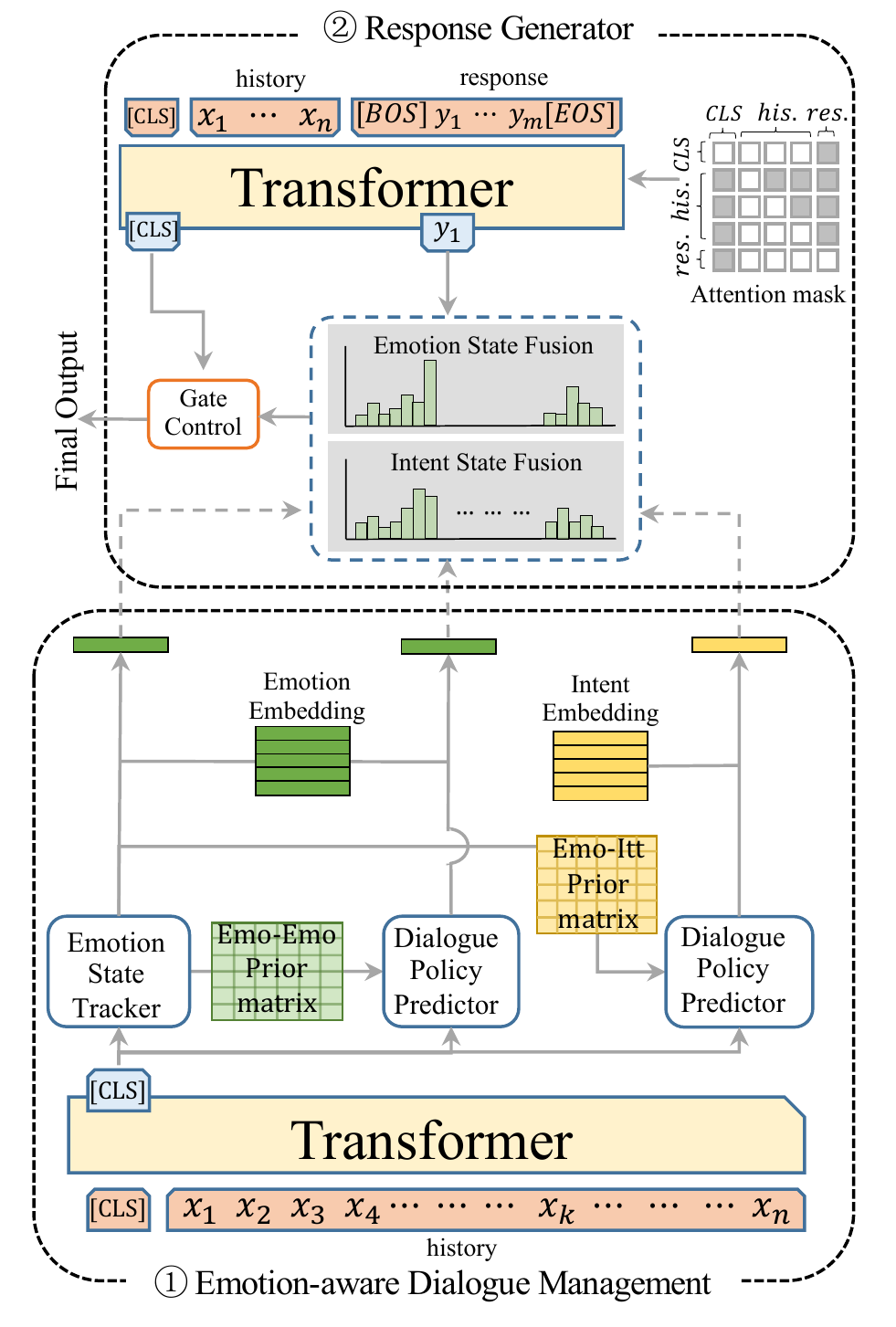}
\caption{Architecture of the proposed model. It contains two parts: (1)Emotion-aware Dialogue Management and (2) Response Generator. Given an input dialogue history, the emotion-aware dialogue management first tracks the speaker's emotion, and then predict listener's emotion and intent based on the result of emotion state tracking module and our proposed emotion shift patterns. The predicted information is then used by the response generator to guide the generation of responses.}
\label{fig:model}
\end{figure}

\subsection{Preliminary}
Given a dialogue history $\{s_0,l_0,s_1,l_1,\cdots,s_M\}$ where $s_*$ and $l_*$ represent utterances from speakers and listeners, the task goal is to generate proper empathetic response $y=\{y_1,\cdots,y_m\}$. Following previous works \cite{lin-etal-2019-moel,rashkin-etal-2019-towards}, we flatten the dialogue history into a sequence $x=\{x_1,\cdots,x_n\}$. We assume that the whole generation process is controlled by speaker emotion $\bm{v}_{\varepsilon_s}$, listener emotion $\bm{v}_{\varepsilon_l}$, and listener intent $\bm{v}_{\tau}$. 
Therefore, the problem can be optimized through maximizing the likelihood $\textstyle\prod_{i=1}^m\mathcal{P}(y_i|y_{<i};x;\bm{v})$.

Since the original dataset \cite{rashkin-etal-2019-towards} does not contain emotion labels and  intent labels for each utterance, we leverage several external datasets. Specifically, we first obtain an emotion prediction model by fine-tuning the BERT model \cite{devlin-etal-2019-bert} on a fine-grained emotion\footnote{the 7 emotion categories are: anger, disgust, fear, joy, neutral, sadness and surprise.} classification dataset, \texttt{goEmotion} \cite{demszky-etal-2020-goemotions}. The emotion prediction model is then used to annotate an emotion label for each utterance. The emotion prediction model achieved an accuracy of 70\% on test set, indicating that it is reliable for emotion classification. The intent\footnote{The 9 intent labels are: questioning, acknowledging, agreeing, consoling, encouraging, sympathizing, wishing, suggesting and neutral} label of each utterance is obtained using the data proposed by \cite{welivita-pu-2020-taxonomy}, where each utterance is annotated with intent information.

As shown in Fig~\ref{fig:model}, our proposed model contains two parts: the Emotion-aware Dialogue Management (EmoDM) is used to predict the overall states and extract their features and a Response Generator (RespG) is used to generate responses conditioned on dialogue context and the states.

\subsection{Emotion-aware Dialogue Management}
\label{subsec: stat-manage}
We build the Emotion-aware Dialogue Management~(EmoDM) on a transformer \cite{vaswani2017attention} encoder. Formally, given a dialogue history sequence $x=\{[\mathrm{CLS}], x_1,\cdots,x_n\}$, the input representation for each word is the sum of its token embedding and position embedding, then they are fed into the transformer encoder to get a sequence of hidden outputs $\{\bm{h_{[\mathrm{CLS}]},h_{x_1},\cdots,h_{x_n}}\}$.

\noindent
\textbf{\textit{Emotion State Tracker}.}
We use the representation of [CLS] token as the context representation of the dialogue history. The representation is then fed into a linear layer to obtain the speaker emotion probability distribution $\mathcal{P}(\varepsilon_s|\bm{x}) = \mathrm{softmax}(\bm{h_{[\mathrm{CLS}]}} \bm{V}_{\varepsilon}^T)$. Specifically, the layer weights are shared with emotion state embedding $\bm{V}_{\varepsilon} \in \mathbb{R}^{N_{\varepsilon}\times d}$. Here $N_{\varepsilon}$ is the number of emotion and $d$ is the vector dimension.

\noindent
\textbf{\textit{Dialogue Policy Predictor}.}
There is strong evidence that emotion-emotion (emo-emo) shift pattern exists among conversations.  We can use this prior knowledge to predict emotion as accurately as possible with only dialogue history context. Through normalizing the frequency that emotion $i$ shifts to $j$, we can compute the shift probabilities to build emo-emo prior matrix $\bm{\mathcal{M}}_{\varepsilon}=[a_{i,j}]\in \mathbb{R}^{N_{\varepsilon}\times N_{\varepsilon}}$ based on training and validation datasets. We can use the predicted speaker emotion to infer some clues about possible shift probabilities. According to the predicted label index $k$,
we can get the whole possible emotion shift probabilities $\bm{m}_{sft}=[a_{k,1},\cdots,a_{k, N_{\varepsilon}}]$. We multiply this prior distribution with the emotion state embedding $\bm{V}_{\varepsilon}$ to get the representation $\bm{v}_{sft} = \bm{V}_{\varepsilon}^T \bm{m}_{sft}$.
Then we can obtain future emotion probability distribution 
\begin{equation}
    \mathcal{P}(\varepsilon_l|\bm{x},\varepsilon_s,\bm{\mathcal{M}}_\varepsilon) = {\rm softmax}(\bm{W_1}[\bm{h_{[\mathrm{CLS}]}};\bm{v}_{sft}] + \bm{b_1})
\end{equation}
Here $\bm{W_1}$ and $\bm{b_1}$ are trainable parameters.

To predict the listener's intent, it is also necessary to use the emotion-intent (emo-itt) shift pattern as the prior knowledge. Similar to listener emotion prediction, we can build the emot-itt prior matrix $\bm{\mathcal{M}}_{\tau}=[b_{i,j}]\in \mathbb{R}^{N_{\varepsilon}\times N_{\tau}}$. Here $N_{\tau}$ is the number of intent.
After that, we can get emotion-intent shift probabilities $\bm{m}_{sft}=[b_{k,1},...,b_{k, N_{\tau}}]$. We use $\bm{m}_{sft}$ an intent embedding $\bm{V}_{\tau} \in \mathbb{R}^{N_{\tau}\times d}$ to get the prior intent $\bm{v}_{sft}= \bm{V}_{\tau}^T \bm{m}_{sft}$  , which is similar to listener emotion prediction. Then we can predict intent probability distribution
\begin{equation}
    \mathcal{P}(\tau|\bm{x},\varepsilon_s, \bm{\mathcal{M}}_\tau) = \sigma(\bm{W_2}[\bm{h_{[\mathrm{CLS}]}};\bm{v}_{sft}] + \bm{b_2})
\end{equation}

Here we use sigmoid instead of softmax since there exist multiple intent labels in one response. $\bm{W_2}$ and $\bm{b_2}$ are trainable parameters.

\noindent
\textbf{\textit{Optimization}.}
The EmoDM can be optimized through a negative log likelihood (NLL) loss, where $\alpha$ and $\beta$ are hyper-parameters:
\begin{equation}
\setlength{\abovedisplayskip}{3pt}
\setlength{\belowdisplayskip}{3pt}
    \begin{split}
        \mathcal{L}_U &= \alpha\log \mathcal{P}(\varepsilon_s|\bm{x}) + (1-\alpha)\log \mathcal{P}(\varepsilon_l|\bm{x},\varepsilon_s,\bm{\mathcal{M}}_\varepsilon) \\ 
        &+ \beta\log \mathcal{P}(\tau|\bm{x},\varepsilon_s, \bm{\mathcal{M}}_\tau)
    \end{split}
\end{equation}

\subsection{Response Generator}
\label{subsec: resp-gen}
Our Response Generator (RespG) is based on a transformer decoder, which is used to generate empathetic responses given a dialogue context and the predicted states.
We concatenate the special token$\mathrm{[CLS]}$, dialogue context $\{x_1, x_2,\cdots,x_n\}$, and ground-truth response $\{y_1, y_2,\cdots,y_m\}$ together and obtain their hidden outputs $\{\bm{h_{[\mathrm{CLS}]},h_{x_1},\cdots,h_{y_m}}\}$ through the transformer decoder. In our RespG, we modify the original attention mask in the original transformer decoder to let $\mathrm{[CLS]}$ attend to the rightward history context for later usage.

During decoding, we fuse the hidden output $\bm{h}_{y_i}$ of $i$-th step with the state feature predicted by EmoDM to control the generation process.  Specifically, we first obtain speaker emotion state $\bm{v}_{\varepsilon_s}$ and listener emotion state $\bm{v}_{\varepsilon_l}$ from $\bm{V}_{\varepsilon}$ according to predicted emotion labels where we can get them from $\mathcal{P}(\varepsilon_s|\bm{x})$ and $\mathcal{P}(\varepsilon_l|\bm{x},\varepsilon_s,\bm{\mathcal{M}}_\varepsilon)$ by argmax operation. Then we use the following approach to inject emotion state bias, where $\bm{W_3}$ and
$\bm{W_4}$ are trainable parameters:
\begin{equation}
\bm{h}_{y_i}^{\varepsilon} = \bm{W_3}(\bm{h}_{y_i} + \bm{v}_{\varepsilon_s}) + \tanh(\bm{W_4}\bm{h}_{y_i}) \bm{v}_{\varepsilon_l}
\end{equation}

Intent prediction is a multi-label classification task, which means the response may have multiple intent labels. 
To fuse multiple intents, we first obtain the intent $\bm{v}_{\tau}$ from $\bm{V}_{\tau}$ with an average operation as in Eq.\ref{eq:avg} and then inject intent features into token hidden outputs to provide intent information:

\begin{align}
\bm{v}_{\tau} &= \hat{I}_{\tau}\bm{V}_{\tau} / {\textstyle \sum_{i=1}^{N_\tau} \hat{I}_{\tau}^{(i)}} \label{eq:avg}\\
\bm{h}_{y_i}^{\tau} &= \bm{v}_{\tau} \odot \bm{h}_{y_i} + \bm{v}_{\tau} \label{eq:h-tau}
\end{align}

\noindent
Where $\odot$ denotes element-wise multiplication and $\hat{I}_{\tau}$ is the predicted multi-label vector. We can get it through applying argmax operation to each row of the intent probability distribution $\mathcal{P}(\tau|\bm{x},\varepsilon_s, \bm{\mathcal{M}}_\tau)$.

Now that two different kinds of state fused representations have been obtained, they are projected into vocabulary logit space. To effectively merge two distributions, we design a gate control layer to monitor the information flow. We pass $\mathrm{[CLS]}$ hidden output through a linear layer with sigmoid activation to obtain the control value $\gamma = \sigma(\bm{W_5} \bm{h}_{\mathrm{[CLS]}} + \bm{b_5})$. After that, we apply softmax to the merged logits to get final probabilities:
\begin{equation}
\setlength{\abovedisplayskip}{3pt}
\setlength{\belowdisplayskip}{3pt}
\mathcal{P}(y_i|y_{<i};x;\bm{v}) = \mathrm{softmax}(\bm{E}_w^T(\gamma \bm{h}_{y_i}^{\varepsilon} + (1-\gamma)  \bm{h}_{y_i}^{\tau})) 
\end{equation}
Where $\bm{W_5}$ and $\bm{b_5}$ are trainable parameters and $\bm{E}_w \in \mathbb{R}^{V \times d}$ is the token embedding lookup table. For optimizing the RespG, we also adopt NLL loss for it.

\begin{table*}[ht!]
	\centering
	\small
	\begin{tabular}{lcccccc} 
		\toprule[1.5pt]
		            & $P_{\rm BERT}$(\%)           & $R_{\rm BERT}$ (\%)          & $F_{\rm BERT}$(\%)                                        & BLEU-4                                                     & DIST-1                                                     & DIST-2                                                      \\ 
		\midrule[1pt]
		MoEL          & 14.5$\pm${\footnotesize 0.9}          & 12.1$\pm${\footnotesize 0.4}          & 13.3$\pm${\footnotesize 0.3}                                       & 1.592$\pm${\footnotesize 0.090}                           & 0.008$\pm${\footnotesize 0.001}                           & 0.076$\pm${\footnotesize 0.005}                            \\
		MIME          &14.9$\pm${\footnotesize 0.6}          & 12.5$\pm${\footnotesize 0.6}          & 13.8$\pm${\footnotesize 0.4}                                       & 1.658$\pm${\footnotesize 0.064}                           & 0.008$\pm${\footnotesize 0.001}                           & 0.055$\pm${\footnotesize 0.004}                            \\
		EmpDG         & 11.3$\pm${\footnotesize 0.4}          & 10.3$\pm${\footnotesize 1.4}          & 10.9$\pm${\footnotesize 0.8}                                       & 1.318$\pm${\footnotesize 0.021}                           & 0.010$\pm${\footnotesize 0.002}                           & 0.072$\pm${\footnotesize 0.018}                            \\
		MultiGPT      & \uline{18.9}$\pm${\footnotesize 0.2}  & \uline{15.5}$\pm${\footnotesize 0.2}  & \uline{17.0}$\pm${\footnotesize 0.1}                               & 1.448$\pm${\footnotesize 0.012}                           & \textbf{0.034}$\pm${\footnotesize 0.001} & \textbf{0.210}$\pm${\footnotesize 0.002}  \\
		\midrule[1pt]
		Ours(Trs) & 15.3$\pm${\footnotesize 0.2}          & 12.7$\pm${\footnotesize 0.1}          & 14.6$\pm${\footnotesize 0.1}                                       & \textbf{1.730}$\pm${\footnotesize 0.141} & 0.017$\pm${\footnotesize 0.002}                           & 0.107$\pm${\footnotesize 0.008}                            \\
		Ours(LM)  & \textbf{19.2}$\pm${\footnotesize 0.2} & \textbf{15.8}$\pm${\footnotesize 0.3} & \textbf{17.6}$\textsuperscript{\textdagger}\pm${\footnotesize 0.2} & \uline{1.666}$\pm${\footnotesize 0.048}  & \uline{0.028}$\pm${\footnotesize 0.001}  & \uline{0.193}$\pm${\footnotesize 0.005}   \\
		\bottomrule[1.5pt]
	\end{tabular}
	\caption{Automatic metric results. we repeat 5 runs with different seeds and average the results for each method. Standard deviations are given in the small text. \textdagger means the results are statistically significant at $p < 0.05$. The bold numbers denote the best results and the underlined numbers denote the second best results.}
	\label{tb:main-ret-state-manage-part1}
\end{table*}

\section{Experiments}
\subsection{Experiment Setup}
We conduct experiments on \textit{Empathetic Dialogue} dataset \citep{rashkin-etal-2019-towards}, which consists of 25k one-to-one open-domain conversations with 8:1:1 train/validation/test split. 

We explore our models with two different initialization methods: the vanilla transformer and the pretrained language model, denoted as \textbf{Ours(Trs)} and \textbf{Ours(LM)} respectively. 
We first use the dataset to fine-tune the RespG backbone to learn some semantic features. Then we alternatively train the EmoDM and the RespG so that RespG can gradually adapt to the influence of the state feature. For Ours(Trs), we train the model for 20 epochs, where the batch size and learning rate are set to 16 and 2e-4. For Ours(LM), we use BERT-base as EmoDM backbone and GPT2-small as ResG backbone. We train the model for 10 epochs, where the batch size and learning rate are set to 16 and 5e-5. The value for $\alpha$ and $\beta$ are 0.6 and 0.5. We use AdamW \citep{loshchilov2018decoupled} as the optimizer.
and apply the schedule sampling \citep{bengio2015scheduled} so that RespG can access true state feature with some probability.
We choose the model where RespG performs best in the validation set for final evaluation. And top-k sampling \citep{Holtzman2020The} is adopted during inference.

\subsection{Evaluation Metrics}
\noindent
\textbf{\textit{Automatic Metrics}.}
Three kinds of automatic metrics are chosen: (1) BLEU: Following \citep{li-etal-2020-empdg,majumder-etal-2020-mime,lin-etal-2019-moel}, we use
BLEU-4 \citep{papineni-etal-2002-bleu} to calculate n-gram overlap ratio. 
2) BERTscore \citep{bert-score} shows stronger system-level and segment-level correlations with human judgments. It uses contextual embeddings of reference and the generated sentence to compute a weighted cosine similarity. We use matching precision, recall and F1 score ($R_{\textup{BERT}}$, $P_{\textup{BERT}}$ and $F_{\textup{BERT}}$) to measure the similarity.
3) Diversity metrics \citep{li-etal-2016-diversity} aims at computing the proportion of different grams in the text to obtain text diversity. We use Dist-\{1,2\} to get diversity score. 

\noindent
\textbf{\textit{Human Judgements}.}
Two types of human evaluations are carried out to fully verify advantages and shortcomings of different models: human rating and human A/B test. Fleiss-Kappa is calculated after annotation to measure the agreement. For Human rating, we randomly sample 100 dialogues and corresponding generated responses of different models. 5 research students are required to give each response a rating score from  \textit{Relevance} aspect, \textit{Fluency} aspect, and \textit{Empathy} aspect (\textit{Rel.}, \textit{Flu.}, and \textit{Emp.}). The score is a 5-point scale (1: not at all, 3: somewhat, 5: very much). 

For the human A/B test, we rearrange the samples in an A-vs-B format, where A is our model and B is another baseline. Another 3 research students are asked to choose the better response for each instance. they can also choose a \textit{Tie} if both are good or bad. 
In order to  keep the anonymization of compared methods, the response order in each sample is totally shuffled and each group of A/B test uses a distinct dialogue context. Annotators are not allowed to discuss with others during evaluation.

\subsection{Baselines}
We compare our models with the following baselines\footnote{To ensure rationality and fairness during comparison, all the baselines use top-k sampling strategy.}: (1) \textbf{MoEL} \citep{lin-etal-2019-moel}: the transformer-based seq2seq \citep{NIPS2014_a14ac55a} model which softly combines the different decoder outputs according to emotion distribution. (2) \textbf{MIME} \citep{majumder-etal-2020-mime}: An extension of the transformer model, which introduces emotion grouping and sampling stochastically to generate responses. (3) \textbf{EmpDG} \citep{li-etal-2020-empdg}: an adversarial model which adopts multi-resolution emotion perception and two discriminators for interacting with the user feedback. 
(4) \textbf{MultiGPT}: similar to \citep{CAiRE2019,zandie2020emptransfo}, we re-implement a multi-task framework with GPT2 as backbone, which is responsible for language modelling, speaker emotion prediction, and listener emotion prediction simultaneously. 

Ablation study is also conducted to thoroughly analyze the impact of different components in our proposed method. The results and analysis can be found in Sec.~\ref{sec:ablation}.

\subsection{Automatic Evaluation Results}

Table \ref{table: auto-rating} illustrates the whole automatic metric results of different models. Comparing to other baselines,
Ours(LM) and Ours(Trs) reaches the highest results in most metrics, indicating that our proposed methods can effectively learn expression patterns in the corpus. 
For Ours(LM), although Dist-1 of ours are slightly worse than MultiGPT, all three BERTscore of ours are higher than other baselines. This indicates that the generated responses of our model have better relevance to history contexts and MultiGPT may generate some noise and unrelated words, which verifies that incorporating with state features is helpful to capture the semantic characteristics accurately. 
Besides, For Ours(Trs), we achieve the best performances in all three metrics among vanilla transformer backbone based methods. Performance of both Ours(Trs) and Ours(LM) illustrate that our proposed method can generate more informative and empathetic responses.

\subsection{Human Evaluation Results}
The results of human ratings and A/B test are illustrated in Table \ref{table: auto-rating} and Table \ref{tb:AB-state-manage}. The Fleiss-Kappa of each three rating aspects and A/B test are 0.29, 0.14, 0.21, and 0.35, which indicate that the multiple annotators almost reach a fair agreement. The results indicate that Ours(LM) obtains the highest rating score in all three rating aspects. In particular, our model exceeds MultiGPT about 0.5 points in both \textit{Relevance} and \textit{Empathy}, which shows the effective control from the overall states. For Ours(Trs), although the model performance is slightly worse than MultiGPT, we still achieve the best results comparing to other transformer based models. 
The human A/B test results also validate the conclusions mentioned above. Due to limited semantic mapping, Ours(Trs) model is not as good as Ours(LM) model. And the high winning percentage of Ours(LM) correlates with high human rating scores. Overall, the two evaluations together manifest that responses from our proposed model are more preferable and of better quality.

\begin{table}[htbp!]
	\centering
	\begin{tabular}{lccc} 
		\toprule[1.5pt]
		              & \textit{Rel.}    & \textit{Flu.}   & \textit{Emp.}     \\ 
		\midrule[1pt]
		MoEL    & 2.70           & 3.81          & 2.62  \\
		MIME    & 2.65           & 3.89          & 2.73  \\
		EmpDG   & 2.74           & 3.71          & 2.65  \\
		MultiGPT & \underline{3.20}           & \underline{4.14}          & \underline{3.04}  \\
		\midrule[1pt]
		Ours(Trs)    & 2.87           & 3.82          & 2.94   \\
		Ours(LM)     & \textbf{3.70} & \textbf{4.37} & \textbf{3.67}  \\
		\bottomrule[1.5pt]
	\end{tabular}
	\caption{Results on human judgement.}
	\label{table: auto-rating}
\end{table}

\begin{table}[htbp!]
	\centering
	\small
	\begin{tabular}{lccc} 
		\toprule[1.5pt]
		              & Win(\%)    & Loss(\%)   & Tie(\%)     \\ 
		\midrule[1pt]
		 Ours(Trs) vs EmpDG    & 45.2 & 29.2 & 25.6  \\
		 Ours(Trs) vs MoEL     & 42.4 & 35.2 & 22.4  \\
		 Ours(Trs) vs MIME     & 43.6 & 34.0 & 22.4  \\
		 Ours(Trs) vs MultiGPT & 25.9 & 50.0 & 24.1  \\
		 \midrule[1pt]
		 Ours(LM) vs EmpDG    & 72.2 & 24.1 & 3.7   \\
		 Ours(LM) vs MoEL     & 77.7 & 11.9 & 10.4  \\
		 Ours(LM) vs MIME     & 61.1 & 25.9 & 13.0  \\
		 Ours(LM) vs MultiGPT & 51.8 & 35.1 & 13.1  \\ 
		 \midrule[1pt]
		 Ours(LM) vs Ours(Trs) & 65.7 & 23.5 & 13.8  \\
		\bottomrule[1.5pt]
	\end{tabular}
    \caption{Results on human A/B test.}
	\label{tb:AB-state-manage}
\end{table}

\begin{table*}[htbp!]
	\centering
	\small
	\begin{tabular}{lcccccc} 
		\toprule[1.5pt]
		                          & $P_{\rm BERT}$(\%)           & $R_{\rm BERT}$(\%)           & $F_{\rm BERT}$(\%)           & BLEU-4                                                     & DIST-1                                                     & DIST-2                                                      \\ 
		\midrule[1pt]
		Ours(Trs)               & \textbf{15.3}$\pm${\footnotesize 0.2} & 12.7$\pm${\footnotesize 0.1}          & \textbf{14.6}$\pm${\footnotesize 0.1} & \textbf{1.730}$\pm${\footnotesize 0.141} & 0.017$\pm${\footnotesize 0.002}                           & 0.107$\pm${\footnotesize 0.008}                            \\
		~ vs. Naive & 15.3$\pm${\footnotesize 0.2}          & \textbf{12.6}$\pm${\footnotesize 0.3} & 14.2$\pm${\footnotesize 0.2}          & 1.521$\pm${\footnotesize 0.074}                           & 0.021$\pm${\footnotesize 0.002}                           & 0.114$\pm${\footnotesize 0.009}                            \\
		~ w/o itt   & 14.6$\pm${\footnotesize 0.3}          & 12.3$\pm${\footnotesize 0.3}          & 13.7$\pm${\footnotesize 0.3}          & 1.486$\pm${\footnotesize 0.090}                           & \textbf{0.028}$\pm${\footnotesize 0.005} & \textbf{0.138}$\pm${\footnotesize 0.024}  \\
		~ w/o emo   & 14.5$\pm${\footnotesize 0.7}          & 12.1$\pm${\footnotesize 0.6}          & 12.4$\pm${\footnotesize 0.3}          & 1.415$\pm${\footnotesize 0.001}                           & 0.018$\pm${\footnotesize 0.003}                           & 0.101$\pm${\footnotesize 0.011}                            \\
		\midrule[1pt]
		Ours(LM)                & \textbf{19.2}$\pm${\footnotesize 0.2} & \textbf{15.8}$\pm${\footnotesize 0.3} & \textbf{17.5}$\pm${\footnotesize 0.2} & \textbf{1.666}$\pm${\footnotesize 0.048} & 0.028$\pm${\footnotesize 0.001}                           & 0.193$\pm${\footnotesize 0.005}                            \\
		~ vs. Naive & 16.7$\pm${\footnotesize 0.5}         & 14.7$\pm${\footnotesize 0.3}          & 15.8$\pm${\footnotesize 0.3}          & 1.149$\pm${\footnotesize 0.181}                           & \textbf{0.305}$\pm${\footnotesize 0.001} & \textbf{0.226}$\pm${\footnotesize 0.005}  \\
		~ w/o itt   & 19.0$\pm${\footnotesize 0.1}          & 15.8$\pm${\footnotesize 0.3}          & \textbf{17.5}$\pm${\footnotesize 0.1} & 1.620$\pm${\footnotesize 0.070}                           & 0.264$\pm${\footnotesize 0.002}                           & 0.182$\pm${\footnotesize 0.010}                            \\
		~ w/o emo   & 18.3$\pm${\footnotesize 0.2}          & 14.8$\pm${\footnotesize 0.3}         & 16.6$\pm${\footnotesize 0.2}          & 1.335$\pm${\footnotesize 0.117}                           & 0.018$\pm${\footnotesize 0.002}                           & 0.140$\pm${\footnotesize 0.012}                            \\
		\bottomrule[1.5pt]
	\end{tabular}
	\caption{Results on Ablation Study. We thoroughly analyze the impact of different components in our method: (1) \textbf{w/o emo}: only the intent is predicted and fused. (2) \textbf{w/o itt}: only emotion state is predicted and fused. (3) \textbf{Naive}: only speaker emotion is predicted, and simply utilize prior shift matrix to get the other states for generation.}
	\label{tb:ablation-study-state-manage-part1}
\end{table*}

\subsection{Ablation Study}
\label{sec:ablation}
Ablation study results are presented in Table \ref{tb:ablation-study-state-manage-part1}.
For Ours(LM), performance of w/o-itt (only utilizing emotion state) is better than that of w/o-emo (only utilizing intent information). Its DIST and $F_{\mathrm{BERT}}$ are higher than those of w/o-emo. which may indicate that emotion states can lead to generating more diverse and more semantic-related words. The emotion state may play a more crucial role in managing generation. Nevertheless, when we combine two states through gate control, the full model achieves the best metric results. This shows that both state feature can provide useful information from different views. For Ours(Trs), such conclusion can also be inferred. 

Besides, there are some slightly different findings when observing the performance of Naive model. For Ours(Trs)-Naive, its BERTscore is better than other two sub-models. For Ours(LM)-Naive, its BERTscore and BLEU is worse than other two sub-models, while it gets the highest DIST score. It illustrates that for an empathetic response model where its parameters are randomly initialized, the introduction of two state features can better help the model to generate empathetic and semantically relevant responses even if its state prediction is not accurate enough. But for the model where its parameters are initialized by pretrained language model, it generates some unrelated words because of inaccurate state prediction.

	

\subsection{Further Analysis}

\begin{table}[H]
    \setlength\tabcolsep{3pt}
	\centering
	\small
	\label{table-NLU-emo}
	\begin{tabular}{lcccc} 
		\toprule[1.5pt]
		 & \multicolumn{2}{c}{speaker}                         & \multicolumn{2}{c}{listener}                         \\
		\multicolumn{1}{c}{}                    & acc(\%)                      & wF1(\%)                      & acc(\%)                      & wF1(\%)                       \\ 
		\midrule[1pt]
		MultiGPT                                & 87.7$\pm$ 0.5         & 87.6$\pm$0.5         & 52.1$\pm$0.4          & 47.1$\pm$1.4           \\ 
		\midrule[1pt]
		Ours(Trs)                               & 76.5$\pm$0.2          & 76.2$\pm$0.2          & 50.4$\pm$0.2          & 42.6$\pm$0.3           \\
		~vs. Naive                                  & 76.3$\pm$0.3          & 76.0$\pm$0.4          & 43.2$\pm$0.3          & 29.4$\pm$0.1           \\
		~w/o itt                                & 77.0$\pm$0.1          & 76.8$\pm$0.1          & 50.7$\pm$0.2          & 43.1$\pm$0.4           \\
		~w/o emo                                & -                        & -                        & -                        & -                         \\ 
		\midrule[1pt]
		Ours(LM)                                & \textbf{88.5}$\pm$0.4 & \textbf{88.4}$\pm$0.4 & \textbf{52.6}$\pm$0.4 & 45.5$\pm$1.2           \\
		~vs. Naive                                  & 87.1$\pm$1.2          & 87.0$\pm$1.2          & 42.4$\pm$0.3          & 29.7$\pm$0.1           \\
		~w/o itt                                & 88.1$\pm$0.3          & 88.0$\pm$0.3          & 51.7$\pm$0.7          & \textbf{48.0}$\pm$0.6  \\
		~w/o emo                                & -                        & -                        & -                        & -                         \\
		\bottomrule[1.5pt]
	\end{tabular}
	\caption{Emotion state prediction results of the EmoDM. \label{table-NLU-emo}}
\end{table}

\begin{table}[H]
	\centering
	\small
	\begin{tabular}{lccc} 
		\toprule[1.5pt]
		        & HL(\%)                       & miF1(\%)                     & AP(\%)                        \\ 
		\midrule[1pt]
		Ours(Trs) & 15.4$\pm$0.2          & 44.2$\pm$0.1          & 54.9$\pm$0.1           \\
		~vs. Naive    & 15.9$\pm$0.1          & 42.4$\pm$0.1          & 52.6$\pm$0.3           \\
		~w/o itt  & -                        & -                        & -                         \\
		~w/o emo  & 15.3$\pm$0.1          & 44.1$\pm$0.2          & 54.8$\pm$0.2           \\ 
		\midrule[1pt]
		Ours(LM)  & \textbf{15.2}$\pm$0.1 & \textbf{44.8}$\pm$0.1 & \textbf{55.3}$\pm$0.1  \\
		~vs. Naive    & 15.8$\pm$0.1          & 42.4$\pm$0.1          & 52.6$\pm$0.1           \\
		~w/o itt  & -                        & -                        & -                         \\
		~w/o emo   & 15.8$\pm$0.1          & 42.8$\pm$0.4          & 52.8$\pm$0.3           \\
		\bottomrule[1.5pt]
	\end{tabular}
	\caption{Intent prediction results of the EmoDM. \label{table-NLU-ITT}}
\end{table}

\noindent
\textbf{\textit{State Prediction Comparison}.}
Next, we evaluate the state prediction performance of the EmoDM.
The results on state prediction are presented in Table \ref{table-NLU-emo} and Table \ref{table-NLU-ITT}. For emotion state prediction performance, we report accuracy (acc) and weighted F1 (wF1). For intent prediction performance, we report hamming loss (HL), average precision (AP) and micro F1 (miF1). 

As can be seen in the table, Ours(LM) model achieves the highest current emotion wF1, future emotion acc, intent AP and intent miF1. Specifically, for speaker emotion prediction and listener emotion prediction, our proposed model achieves 88.32\% and 52.67\% accuracy (acc). Comparing to MultiGPT which get  87.61\% and 51.81\% acc, we get better results than it. We also achieve 55.60\% averaged precision (AP) in intent prediction, which means our model has a good capability of predicting intents. This demonstrates that removing either intent prediction component or emotion prediction component will hurt the EmoDM performance. And all the metric scores of Ours(LM) are higher than MultiGPT. This shows that the EmoDM in our proposed methods can get better results than adopting the multi-task framework in this task. 

Besides, for Naive method, we also find that the results of Naive method is not as good as our proposed methods. The speaker and listener emotion prediction acc are 87.67\% and 42.69\%, and the intent AP is 52.65\%. This shows that simply using state shift statistics for prediction is not accurate, and our proposed EmoDM is more effective.

	\begin{table*}[htbp!]
		\centering
		\small
		\begin{tabular}{ll} 
			\toprule[1.5pt]
			State     & sadness \textbar{} sadness \textbar{} questioning                                                                                                                       \\
			Context   & my wife has been really sad here recently, i have had to hold her while she cries herself to sleep at night.                                                            \\
			reference & that is very sad. did something recently happen to her?                                                                                                                 \\ 
			\midrule[1pt]
			MoEL      & that is a great thing to have a little bit of the pain .                                                                                                                \\
			MIME      & \textbf{i know that feeling} . how long ago was it for you ?                                                                                                            \\
			EmpDG     & \textbf{oh no !} how did she do that ?                                                                                                                                  \\
			MultiGPT  & \begin{tabular}[c]{@{}l@{}}You can always talk to her about it though! It's important to remember that when you're alone, \\it takes care of themselves! \end{tabular}  \\
			Ours(Trs)  & i am \textbf{so sorry} to hear that . i \textbf{hope} she is \textbf{doing better} now                                                                                  \\
			Ours(LM)   & I'm \textbf{so sorry}. \textbf{I know that feeling}. Is \textbf{she ok? }                                                                                               \\ 
			\hline\hline
			State     & joy \textbar{} surprise \textbar{} acknowledging,questioning                                                                                                            \\
			Context   & I have recently become vegan and I am very impressed with the food                                                                                                      \\
			reference & that is interesting . do you miss meat or dairy at all ?                                                                                                                \\ 
			\midrule[1pt]
			MoEL      & i \textbf{love }my \textbf{food }. i do not think i do not like to have those things like that !                                                                        \\
			MIME      & that is \textbf{great !} i love my family , but it was nice to have a great time                                                                                        \\
			EmpDG     & thats \textbf{awesome }! what kind of i was it                                                                                                                          \\
			MultiGPT  & What type of \textbf{foods }are you getting?                                                                                                                            \\
			Ours(Trs)  & That is \textbf{awesome}. I have never considered \textbf{vegan}, so it is \textbf{awesome}.                                                                            \\
			Ours(LM)   & that is \textbf{awesome }! what kind of \textbf{food }is it ?                                                                                                           \\
			\bottomrule[1.5pt]
		\end{tabular}
		\caption{Example responses generated by our proposed model and other baselines. In State row, $*|*|*$ denotes speaker's emotion, listener's emotion, and listener's intent. \label{table: case-study}}
	\end{table*}

\begin{figure}[htbp!]
\centering
\includegraphics[scale=0.58]{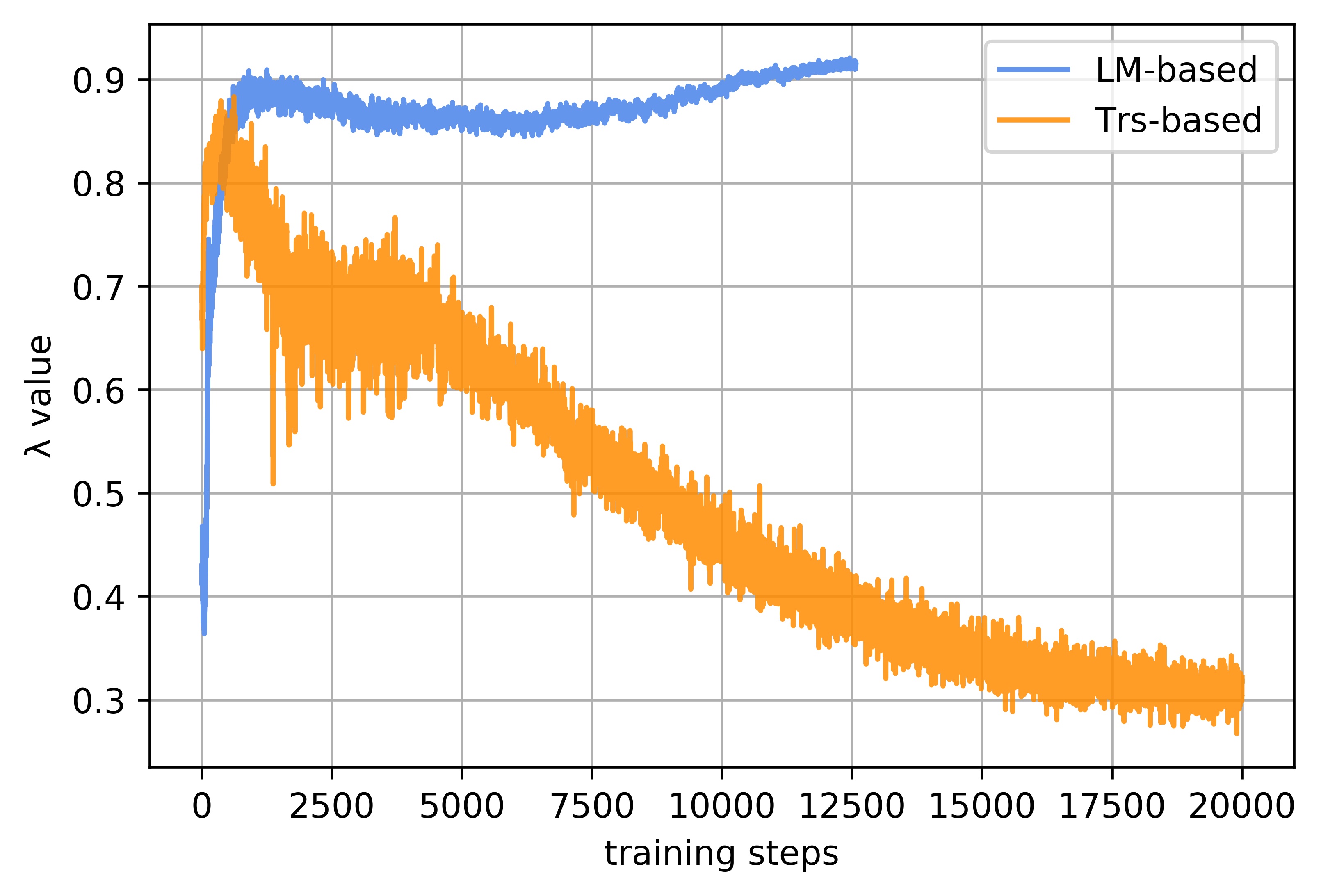}
\caption{Impact of $\lambda$ value on model training. \label{fig:gate_control_value}}
\end{figure}

\noindent
\textbf{\textit{Gate Control}.}
Fig.\ref{fig:gate_control_value} visualizes changing tendencies of $\lambda$ for our two models during training. 
In the test set, for Ours(LM), gate control value $\lambda$ ranges from 0.8536 to 0.8951. And for Ours(Trs), $\lambda$ ranges from 0.2122 to 0.2588. 
For Ours(LM), it seems that more information is provided by emotion states, whereas intents are also indispensable. For Ours(Trs), such a conclusion is opposite. 
We analyze that maybe intent belongs to elementary mental state and is easier to learn, but emotion belongs to advanced mental state \citep{pettinelli2012psychology} and plays a bigger role in controlling generation. 
If the RespG of the model has already got a deep semantic understanding, it will pay more attention to emotion states.

\subsection{Case Study}
As illustrated in Table \ref{table: case-study}, we list two detailed cases generated by our proposed models and other baselines. In the first case, nearly all the generated responses correctly match the history situation, but responses from our proposed models are more informative and easier to resonate with the speaker. Our models not only convey a strong sympathetic feeling, but also illustrate obvious empathy by using acknowledging and questioning propositions like "I know that feeling, is it ok?". In the second case, although the general response emotions of several baselines are approximately correct, our models can also generate some keywords like "vegan" and "food". Besides, responses of our models have fewer logical errors and grammar mistakes under state constraints than those of other models.

\section{Conclusion}
In this paper, we propose a novel empathetic response generation model with emotion-aware dialogue management. The emotion-aware dialogue management consists of two modules: (1) the Emotion State Tracker maintains the current emotion state of the user, and (2) the Empathetic Dialogue Policy Predictor predicts the response emotion and the intent of the response. To model interaction between the use's emotion and the responding strategy, we also introduce emotion shift patterns. Experimental results show that our model outperforms other state-of-the-art baselines in both automatic and human evaluations, and the proposed emotion-aware dialogue management can effectively improve empathetic response generation. 

In the future, it is possible to use some more powerful dialogue management strategies such as reinforcement learning and adding empathy intensity measurement as feedback to further improve the model performance.

\normalem
\bibliography{anthology,custom}




\end{document}